\pdfoutput=1

\documentclass[11pt]{article}

\usepackage{amsmath,amsfonts,bm}

\def\eqref#1{equation~\ref{#1}}

\def\1{\bm{1}}

\DeclareMathAlphabet{\mathsfit}{\encodingdefault}{\sfdefault}{m}{sl}
\SetMathAlphabet{\mathsfit}{bold}{\encodingdefault}{\sfdefault}{bx}{n}

\usepackage{color}
\usepackage{times}
\usepackage{epsfig}
\usepackage{graphicx}
\usepackage{amsmath}
\usepackage{amsthm}
\usepackage{amssymb}
\usepackage{mathtools}
\usepackage{multirow}
\usepackage{bbm}
\usepackage{dsfont}

\usepackage[table, dvipsnames]{xcolor}

\usepackage[utf8]{inputenc}
\usepackage{booktabs,siunitx}
\usepackage{multirow}
\usepackage{multicol}
\usepackage{amsmath}
\usepackage{subcaption}
\usepackage{caption}
\usepackage{graphicx}
\usepackage{pifont}

\usepackage{tikz}
\usetikzlibrary{shapes,arrows,patterns}
\usetikzlibrary{positioning}
\usetikzlibrary{calc}

\newcommand{\cut}[1]{}

\newcommand{\postspace}{\vskip -3mm}
\newcommand{\minipostspace}{\vskip -2mm}

\definecolor{red}{RGB}{255, 117, 115}
\definecolor{green}{RGB}{171, 255, 175}

\usepackage[final]{acl}

\usepackage{times}
\usepackage{latexsym}

\usepackage[T1]{fontenc}

\usepackage[utf8]{inputenc}

\usepackage{microtype}

\usepackage{inconsolata}

\usepackage{graphicx}

\usepackage{graphicx}
\definecolor{cadmiumgreen}{rgb}{0.0, 0.42, 0.24}
\definecolor{cardinal}{rgb}{0.77, 0.12, 0.23}
\definecolor{cadmiumred}{rgb}{0.89, 0.0, 0.13}

\usepackage{tcolorbox}
\newtcolorbox[list inside=prompt,auto counter,number within=section]{prompt}[1][]{
    fontupper=\ttfamily\footnotesize,
    boxsep=5pt,
    left=0pt,
    right=0pt,
    top=0pt,
    bottom=0pt,
    boxrule=1pt,
    #1,
}

\title{Learning Beyond the Surface: How Far Can Continual Pre-Training with LoRA Enhance LLMs' Domain-Specific Insight Learning?}

\author{Pouya Pezeshkpour\\
Megagon Labs\\
\texttt{pouya@megagon.ai} \\
\And
Estevam Hruschka\\
Megagon Labs \\
\texttt{estevam@megagon.ai}}

\begin{document}
\maketitle
\begin{abstract}
Large Language Models (LLMs) have demonstrated remarkable performance on various tasks, yet their ability to extract and internalize deeper insights from domain-specific datasets remains underexplored. In this study, we investigate how continual pre-training can enhance LLMs' capacity for insight learning across three distinct forms: declarative, statistical, and probabilistic insights. Focusing on two critical domains: medicine and finance, we employ LoRA to train LLMs on two existing datasets. To evaluate each insight type, we create benchmarks to measure how well continual pre-training helps models go beyond surface-level knowledge. We also assess the impact of document modification 
on capturing insights. The results show that, while continual pre-training on original documents has a marginal effect, modifying documents to retain only essential information significantly enhances the insight-learning capabilities of LLMs. We released our dataset and code\footnote{\url{https://github.com/megagonlabs/insight_miner}}.
\end{abstract}

\section{Introduction}
Large Language Models (LLMs) have demonstrated extraordinary capabilities across a broad spectrum of NLP tasks, from text generation to reasoning and summarization \citep{touvron2023llama,openai2023gpt-4,team2023gemini}. Despite these advancements, a crucial question persists: To what extent can LLMs internalize and utilize deeper insights from domain-specific datasets? While surface-level patterns and explicit knowledge can often be captured and delivered to LLMs using techniques such as retrieval-augmented generation (RAG) \citep{lewis2020retrieval,gao2023retrieval}, extracting and leveraging deeper insights remains a significant challenge.

Solving complex tasks or answering intricate questions often requires accessing deeply buried information within documents or recognizing patterns distributed across numerous samples. 
Additionally, such knowledge can be ambiguous or context dependent and may not be universally correct outside the scope of domain-specific data, posing significant challenges for existing approaches like RAG. We classify these insights into three categories: \textbf{Declarative} insights representing explicit, factual knowledge directly stated in the dataset, such as definitions, facts, or specific details. 
\textbf{Statistical} insights arise from aggregations, distributions, and quantitative summaries observed across multiple data points. And, \textbf{Probabilistic} insights, involve reasoning under uncertainty, inferring likelihoods, and drawing conclusions from incomplete or ambiguous information. Together, these insight types encompass a range of knowledge necessary for nuanced understanding and problem-solving.

\begin{figure}[t!]
    \centering
    \includegraphics[width=\linewidth]{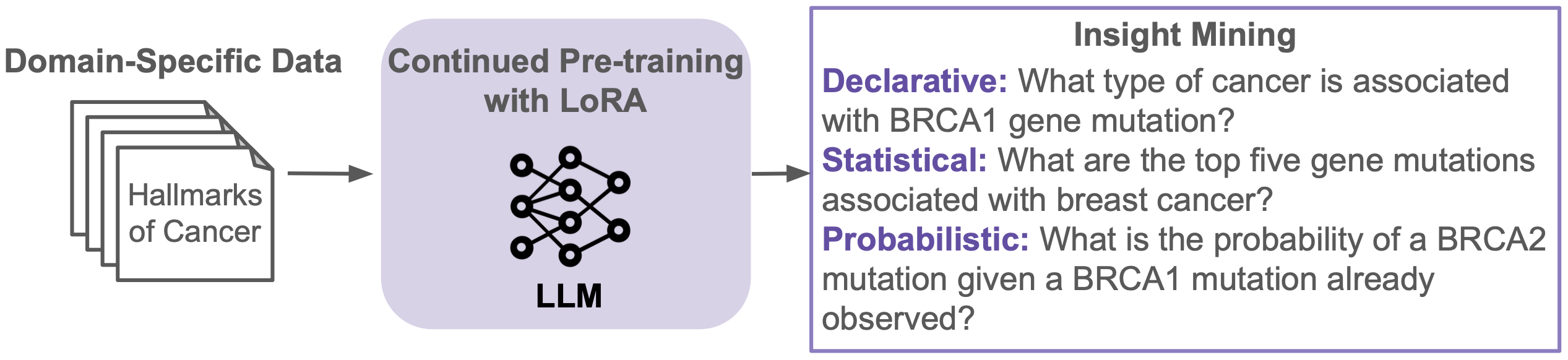}
    \caption{We use domain-specific data, like the Hallmarks of Cancer dataset, to adapt an LLM through continued pre-training with LoRA. Our goal is to assess whether LLMs are capable of effectively capture three types of insights: declarative, statistical, and probabilistic.}
    \label{fig:illust}
    \minipostspace
\end{figure}

Through our investigation, we aim to answer two key research questions:
How effectively can LLMs internalize declarative, statistical, and probabilistic insights through continual pre-training \citep{gururangan2020don,ke2023continual} with low-rank adaptation (LoRA) \citep{hu2021lora}?
And, to what extent can simplifying documents to train LLMs on them enhance their insight-learning capabilities? 
Building on the success of continual pre-training \citep{ke2023adapting,ke2023continual} and LoRA \citep{zhao2024lora} in acquiring new knowledge and solving tasks, our goal is to investigate the capability of LLMs to extract insights and solve tasks in scenarios where tasks and required insights are not predetermined. This motivates our emphasis on continual pre-training, which enables models to adapt dynamically without relying on predefined tasks which is necessary for supervised fine-tuning.

In this work, we address the challenge of enhancing LLMs' capability to learn insights from a domain-specific dataset through continual pre-training using LoRA (Figure \ref{fig:illust}). 
Specifically, we adapt LLaMA-3.2 1B, LLaMA-3.2 3B, and LLaMA-3.1 8B models to two domain-specific datasets: Hallmarks of Cancer \citep{baker2016automatic} for the medicine domain and Buster \cite{zugarini2024buster} for the finance domain. To assess the insight-mining capability of the continued pre-trained LLMs using these datasets, we first 
use GPT-4o mini \cite{hurst2024gpt} to extract triples of information from these datasets, and then manually filter and normalize the relations. Using the final triples, we create an evaluation set for each type of insight.

Our experiments indicate that LLMs with continued pre-training using LoRA exhibit marginal improvements in capturing declarative and statistical insights, with even smaller gains for probabilistic insights. However, training LLMs on modified documents containing only the essential knowledge, represented as triples of information, leads to significant improvements in their insight-learning capabilities. Breaking down LLM performance across different relation types reveals significant variations, emphasizing the potential influence of the models' prior understanding of relations on their ability to acquire new knowledge. Moreover, to further test the limits of LoRA continual pre-training, we additionally trained the models on individual triples instead of the document format. This approach resulted in a substantial improvement in the models' insight-learning capabilities, further emphasizing the critical role of input format in shaping LLMs' ability to learn insights. Across all our experiments, we consistently observed that larger model sizes greatly enhance the capacity to learn new insights, highlighting the scalability of insight learning with increased model capacity.

\section{Data-Specific Insights}
Extracting meaningful insights from large datasets is essential to solve different tasks across domains such as medicine, finance, education, and technology. Insights enable models to make predictions, answer questions, and support evidence-based decisions. For example, in medicine, analyzing patient records and research papers can uncover disease indicators, symptom-outcome correlations, or treatment effectiveness. In finance, insights from market data and financial reports can reveal stock patterns, predict economic trends, and guide investment strategies. In this section, we first define different types of insights investigated in this work, and then describe how we benchmark the insight-mining capability of LLMs across these domains.

\subsection{Insight Types}
Not all insights are of the same nature. They vary in complexity, interpretability, and the reasoning required to uncover them. We classify the insights into three primary types: \textit{Declarative}, \textit{Statistical}, and \textit{Probabilistic}. 

\textbf{Declarative Insights} refer to explicit, factual knowledge directly stated in a dataset. These insights are often presented as concrete pieces of information, such as definitions, facts, or specific details. Most existing works focus on these types of insights, which require minimal inference and are typically retrieved rather than deduced. However, when such insights are deeply buried within documents, they can pose significant challenges for retrieval models.  As an example, in a medical dataset, a declarative insight might state: "\textit{The BRCA1 gene mutation is associated with an increased risk of breast cancer.}" These insights are essential for providing users with accurate and direct answers to factual queries, such as "\textit{What type of cancer is associated with the BRCA1 gene mutation?}"

\textbf{Statistical Insights} emerge from patterns and trends observed across multiple data points. These insights often involve analyzing aggregated data to identify correlations, distributions, and generalizable trends. They require the model to abstract knowledge from repeated observations. In a financial dataset, a statistical insight might reveal: "\textit{The top 3 companies with the highest debt-to-equity ratio are X, Y, and Z.}"

\textbf{Probabilistic Insights} involve reasoning under uncertainty, inferring likelihoods, and drawing conclusions from incomplete or ambiguous information. These insights are crucial in scenarios where definitive answers are not available and predictions must be made based on probabilities. In a medical dataset, a probabilistic insight might suggest: "\textit{Given the patient's symptoms and test results, there is a 70\% chance they have Condition Y.}"

Extracting declarative, statistical, and probabilistic insights from datasets enables LLMs to perform tasks that require factual knowledge, pattern recognition, and reasoning under uncertainty. Each type of insight plays a unique role in enabling models to address real-world challenges effectively. 

\subsection{Benchmarking}
To evaluate the insight-mining capability of continued pre-trained LLMs, we focus on two domain-specific datasets: \textit{Hallmarks of Cancer} \citep{baker2016automatic}---contains research articles, and clinical studies focused on the biological mechanisms underlying cancer---for the medicine domain and \textit{Buster} \citep{zugarini2024buster}---includes financial reports, market analyses, investor statements, and economic forecasts---for the finance domain. These datasets were chosen because they are domain-specific, consist of several thousand documents of varying sizes, and include entities linked through a wide range of relationships, making them ideal candidates for assessing different types of insight extraction. 

To systematically benchmark LLMs, drawing inspiration from previous works \citep{papaluca2023zero, wadhwa2023revisiting}, we first use \textit{GPT-4o mini} to extract \textit{triples of information} in the form of <~\textit{subject-relation-object}~> from the documents (the prompt is provided in the Appendix). These triples are then: (1) \textbf{filtered} to remove irrelevant, noisy, or rare triples ensuring high-quality data, and (2) \textbf{manually normalized}, standardizing the relations to maintain consistency across the dataset. The refined triples, then form the foundation for evaluating the three types of insights. For declarative insights, we focus on subject-relation pairs that have one object, where models are tasked with predicting the object given the subject-relation pair. For statistical insights, we use subject-relation pairs with more than one object, where LLMs are tasked with predicting all objects given the subject-relation pair. For probabilistic insights, we evaluate the capability of LLMs to capture the probability of $\text{entity}_2$ occurring given the presence of $\text{entity}_1$. To construct an evaluation set for this conditional probability, we identify pairs of entities that co-occur within a document and calculate the conditional probability $p(\text{entity}_2 \mid \text{entity}_1)$ over each dataset. These calculated probabilities are then used as queries to assess the model's ability to capture probabilistic insights. It is important to note that in this work we focus on the most atomic form of queries/knowledge. More complex tasks, such as predicting the subject or handling multi-step queries, are left for future research.

To ensure uniformly distributed evaluation sets, we sample 500 queries as evenly as possible. This is achieved by first selecting an equal number of samples from each class, defined by the number of objects in statistical insights and five probability bins in probabilistic insights uniformly covering 0-1 probabilities. Classes without sufficient samples are removed, and the remaining samples are drawn uniformly from the remaining classes. This process continues iteratively until we reach 500 samples. The details of benchmark statistics are provided in the Appendix.

\paragraph{Document Simplification} Since documents often contain a significant amount of unnecessary or distracting information, it can be challenging for LLMs to focus on the most relevant content during training. To mitigate this issue, we also perform continual pre-training of LLMs on a processed version of the documents. In this approach, for each document, we retain only the identified triples of information, which appear in the form of sentences, while discarding all other content. 
By focusing on these triples, we aim to reduce noise and enhance the model’s ability to learn and internalize critical insights, leading to more effective insight extraction across different types.
\begin{figure*}[th!]
    \centering
    \begin{subfigure}[b]{0.32\textwidth}
        \centering
        \includegraphics[width=\linewidth]{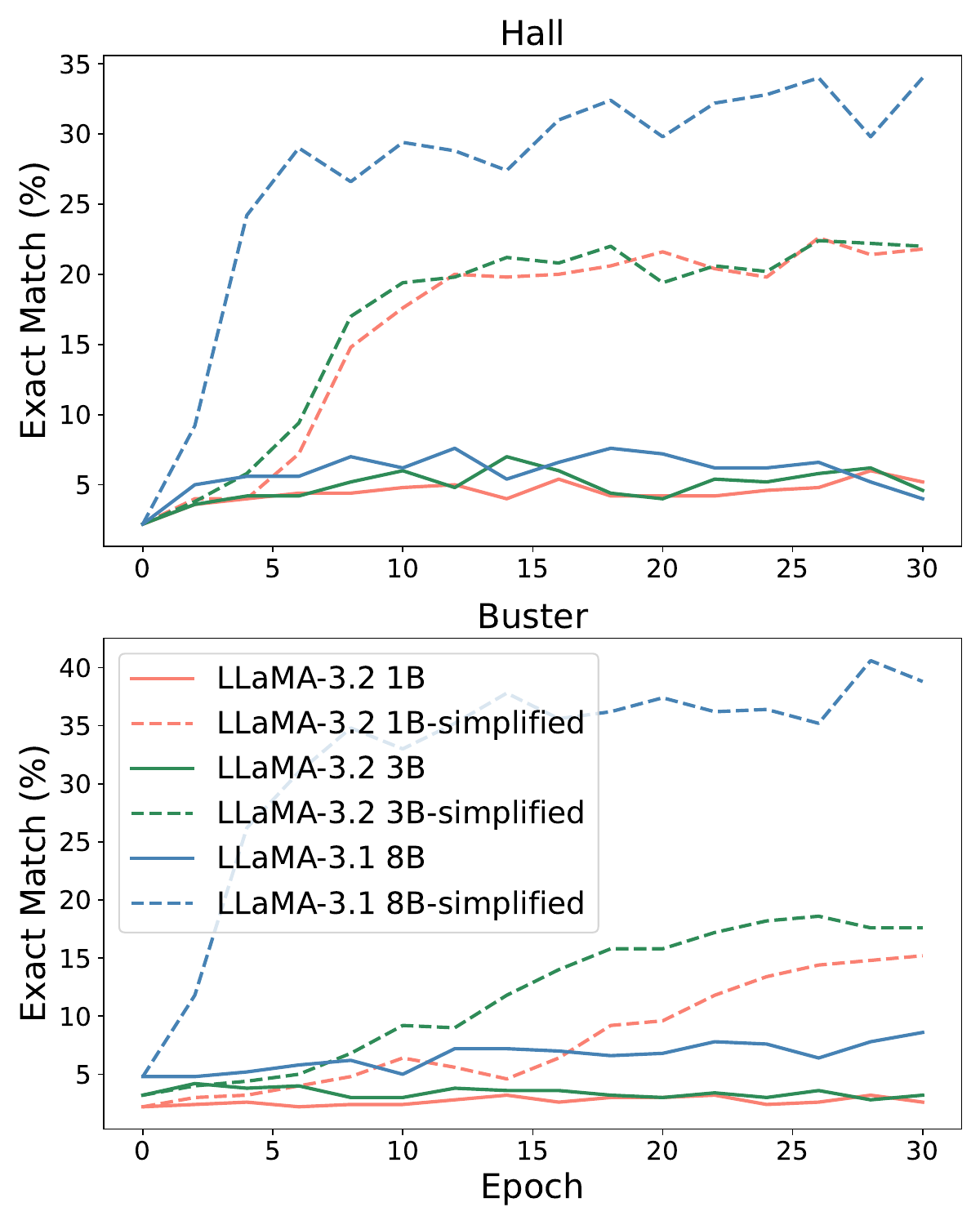}  
        \caption{Declarative}
    \end{subfigure}
    \begin{subfigure}[b]{0.32\textwidth}
        \centering
        \includegraphics[width=\linewidth]{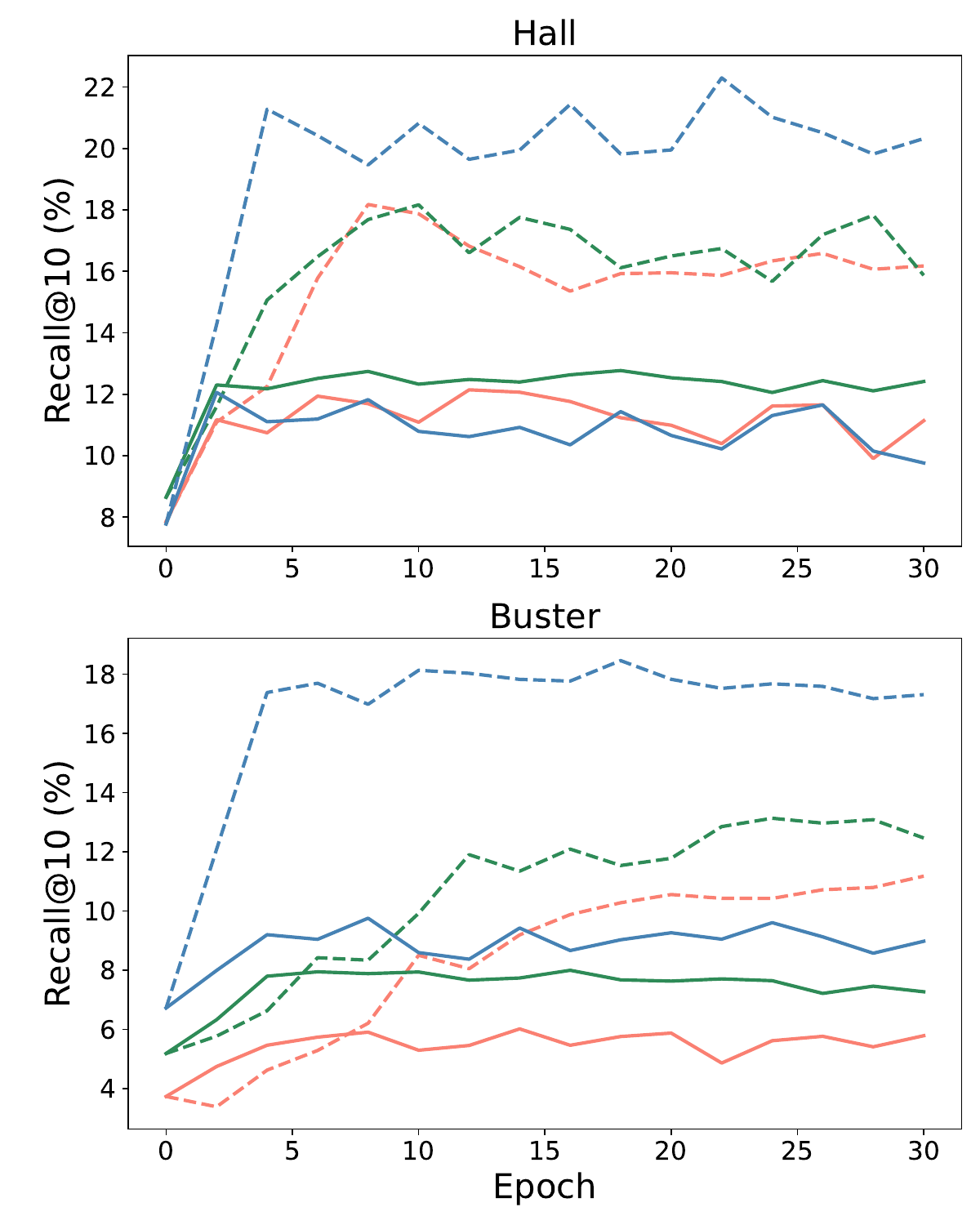}
        \caption{Statistical}
    \end{subfigure}  
    \begin{subfigure}[b]{0.32\textwidth}
        \centering
        \includegraphics[width=\linewidth]{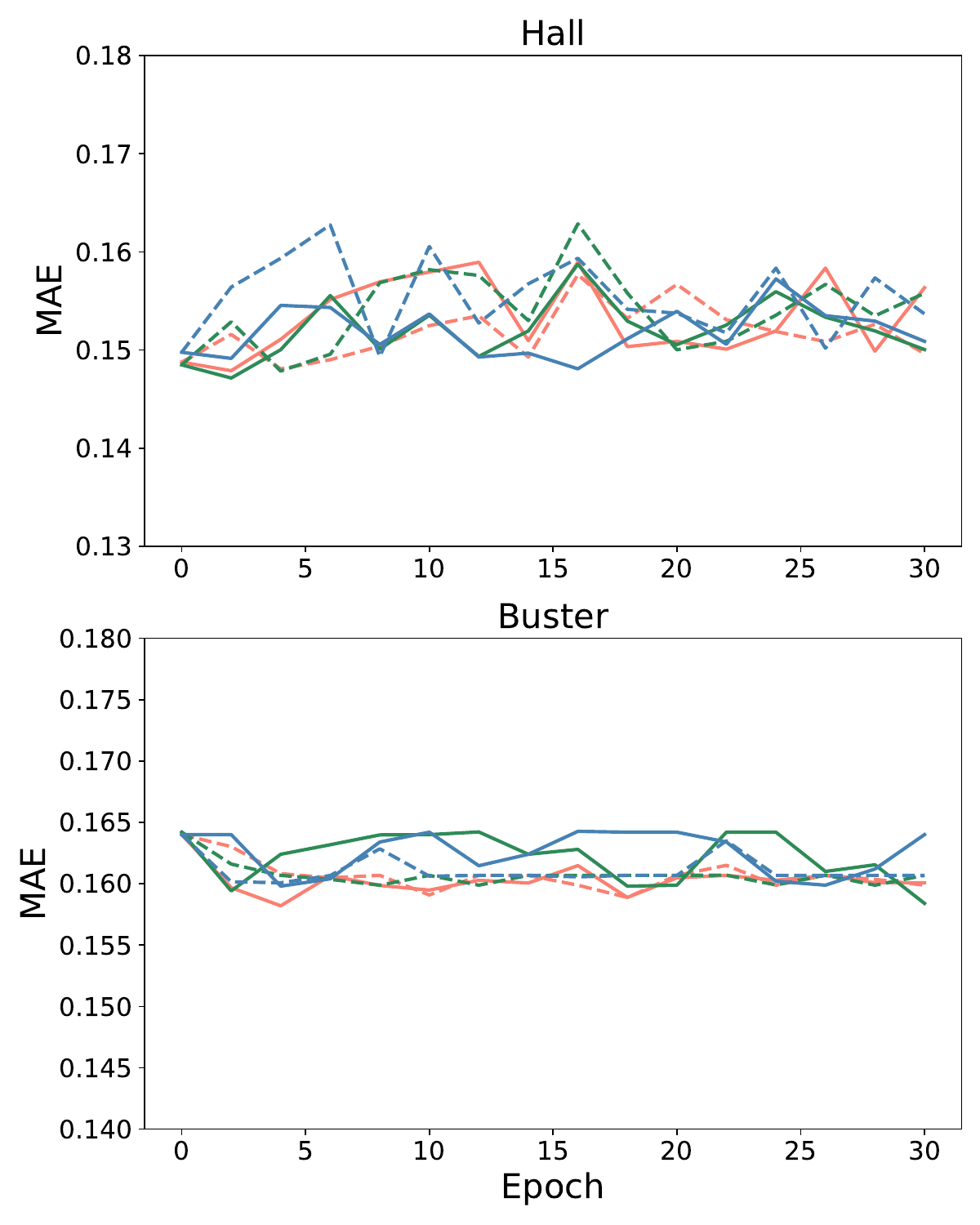}
        \caption{Probabilistic}
    \end{subfigure}  
    \caption{LLMs performance on insight extraction during continual pre-training. Declarative and statistical insights show slight improvement, while probabilistic insights remain largely unchanged. Increasing model size and using simplified documents significantly enhance performance.}
    \label{fig:res}
    \postspace
\end{figure*}



\section{Experimental Details}
\paragraph{Models:} We consider three variants of the LLaMA family of models with various size—LLaMA-3.2 1B, LLaMA-3.2 3B, and LLaMA-3.1 8B—using LoRA to efficiently adapt them to the datasets. We employ continual pre-training to adapt LLMs to the datasets in an unsupervised manner. The motivation for choosing this approach over supervised adaptation is that, in real-world scenarios, the type and exact nature of the insights required might not be clear beforehand. Continual pre-training enables the model to access these insights dynamically without requiring predefined task-specific labels from the data.

\paragraph{Insight Mining} To extract declarative and statistical insights from LLMs, we use the top-1 and top-k predictions from the models, respectively, given the concatenated string of the subject and relation. For probabilistic insights, inspired by previous works on self-consistency \citep{wang2022self,chen2023universal}, we provide only $\text{entity}_1$ to the models, sample top-k generations, and calculate $p(\text{entity}_2\mid\text{entity}_1)$ by counting how often $\textit{entity}_2$ appears in the outputs. It is worth noting that we also explored other methods of calculating probabilities, such as verbally asking the models for a probability or using output logits. However, the sampling approach consistently yielded the best results, aligning with previous work on the confidence approximation \citep{lyu2024calibrating}.

\paragraph{Evaluation Metrics:} We evaluate LLMs performance on declarative insights using two widely adopted metrics in question-answering studies: (1) Exact Match (EM): Measures whether the ground truth appears exactly in the predicted output, and (2) F1 Score: Calculates the average overlap between the gold answer and the predicted response. For statistical insights, we evaluate using Recall@K, which measures the proportion of correct predictions in the top-k predictions relative to the total number of correct answers. Additionally, we use mean absolute error (MAE) and Pearson correlation coefficient to evaluate probabilistic insights.
Further details regarding dataset statistics, hyperparameters, and model configurations are provided in the Appendix.

\begin{table}[t!]
\small
\centering
\begin{tabular}{llrrr}
\toprule 
&&\bf Dec&\bf Stat&\bf Prob\\
\midrule
\multirow{3}{*}{\rotatebox[origin=c]{90}{\bf Hall}}&LLaMA-3.2 1B&98.4&66.7&0.151\\
&LLaMA-3.2 3B&98.8&74.0&\bf0.147\\
&LLaMA-3.1 8B&\bf 99.0&\bf 82.0&0.148\\
\midrule
\multirow{3}{*}{\rotatebox[origin=c]{90}{\bf Buster}}&LLaMA-3.2 1B&97.2&36.1&\bf 0.161\\
&LLaMA-3.2 3B&97.4&43.3&0.162\\
&LLaMA-3.1 8B&\bf 97.8&\bf 50.3&0.162\\
\bottomrule
\end{tabular}
\caption{Impact of continued pre-training of LLMs exclusively on the extracted triples. Performance is reported using Exact Match for declarative insights, Recall@10 for statistical insights, and MAE for probabilistic insights after 30 epochs of pre-training.}
\label{tab:cpt-graph}
\postspace
\end{table}

\section{Experiments}
In this section, our goal is to address two fundamental questions: (1) Can continued pre-training with LoRA 
help LLMs capturing different type of insights? (2) Does processing the original data enhance the model's capability to extract the insights effectively? 

\paragraph{Effectiveness of Continued Pre-Training with LoRA:} The results in Figure \ref{fig:res} demonstrate that continued pre-training marginally improves the models' ability to capture all three types of insight, consistent with previously observed limitations of LoRA \citep{biderman2024lora}. Declarative and statistical insights show slight improvements, with exact match scores increasing by a few percentage points across model sizes. However, probabilistic insights remain largely unchanged and continue to be the most challenging for all models. Moreover, larger models (e.g., LLaMA-3.1 8B) consistently outperform smaller models, highlighting the importance of model capacity in insight learning. We provide F1 scores for declarative, Recall@5 for statistical, and Pearson correlation results for probabilistic insights that further reinforce and emphasize these observations in the Appendix.

\paragraph{Impact of Simplifying the Original Data:}
To evaluate the impact of document simplification on insight learning, we conducted continual pre-training on simplified documents. As shown in Figure \ref{fig:res}, this approach led to noticeable improvements in declarative and statistical insights, with larger models demonstrating greater gains. However, probabilistic insights exhibit similar behavior as before, further emphasizing the challenges in learning this type of insight.

\paragraph{Do LLMs learn insights uniformly across different relations?} 
We provide a per-relation breakdown of LLM performance in capturing declarative and statistical insights for the top five most frequent relations in each dataset, in the Appendix. The results show that, although the top five frequent relations have similar distributions across the datasets, the impact of continual pre-training and the use of original versus simplified documents on LLM insight-mining performance varies significantly across relations. This variation may stem from the differing levels of prior understanding that LLMs possess for these relations.

\paragraph{Do LoRA's limitations in knowledge acquisition hinder LLMs' ability to learn insights, and can performance be improved by further simplifying the information?} 
To address these questions, instead of training models on the document format, we concatenate the components of each triple, treat them as separate inputs, and conduct continual pre-training on this processed format. The results reveal that training on triples enables LLMs to achieve near-perfect performance on declarative insights, highlighting that LoRA's learning capacity is sufficient to capture this level of structured information. In contrast, while performance on statistical insights improves significantly, a considerable gap remains---especially in the Buster dataset---when compared to declarative insights, indicating LLMs' limitations in effectively aggregating information. Furthermore, the lower performance on statistical insights in Buster compared to Hallmarks of Cancer may stem from the higher variety of relations and the larger volume of triples in the Buster dataset. Finally, as observed previously, larger LLMs consistently perform better, while probabilistic insights continue to present similar challenges as in earlier observations.

\section{Conclusion}
We investigate the impact of continual pre-training with LoRA alongside the effect of document processing on LLMs' ability to extract declarative, statistical, and probabilistic insights from domain-specific datasets. Using two existing datasets from the medicine and finance domains, we create benchmarks to evaluate the insight-mining capabilities of LLMs. Our findings reveal that continual pre-training on original documents yields only marginal improvements across all insight types. In contrast, modifying the document format to retain only essential information significantly boosts performance, particularly for declarative and statistical insights. 

\bibliography{main}

\appendix

\section{Experimental Details}

\begin{table*}[th!]
\small
\centering
\begin{tabular}{lrrrr}
\toprule 
\bf Dataset& \bf \# Documents&  \bf \# Extracted Triples &  \bf \# Relations&  \bf \# Entities\\
\midrule
Hallmarks of Cancer&1,852&13,084&135&16,050\\
Buster&9,972&88,078&500&86,641\\
\bottomrule
\end{tabular}
\caption{Data statistics for the Hallmarks of Cancer and Buster datasets, including details of the extracted triples from each.}
\label{tab:stat}
\end{table*}

\begin{figure*}[th]
    \centering
    \begin{subfigure}[b]{0.45\textwidth}
        \centering
        \includegraphics[width=\linewidth]{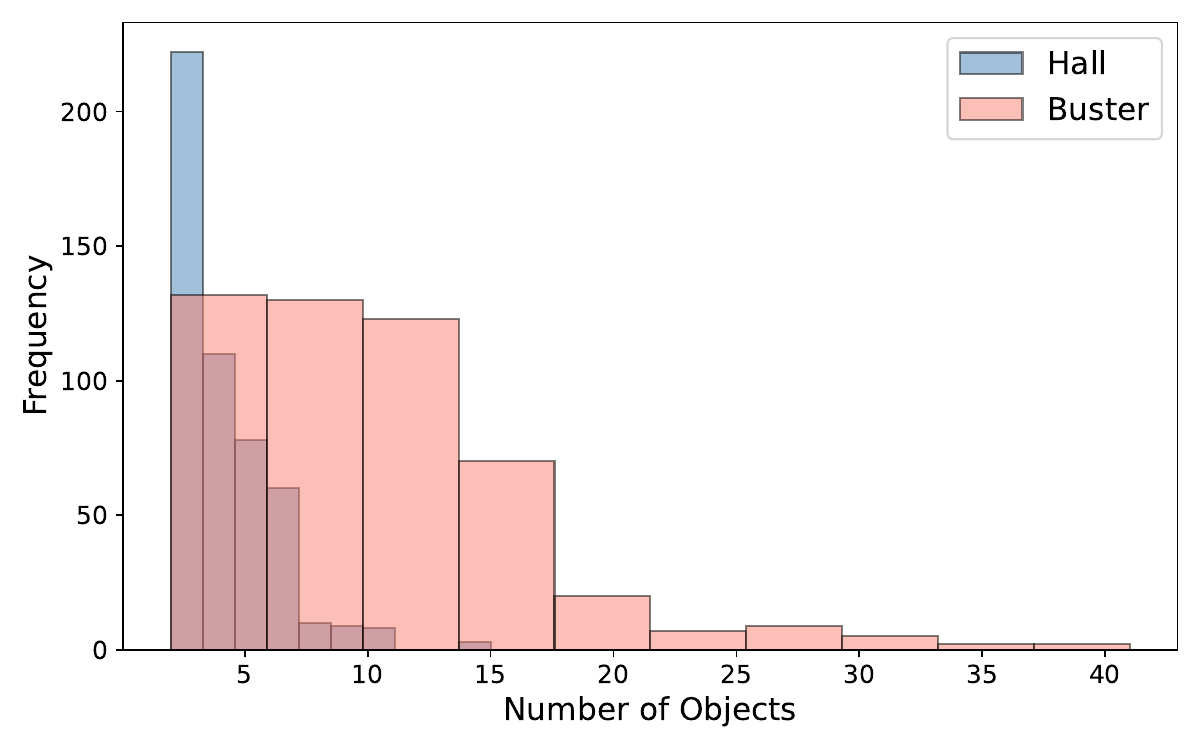}  
        \caption{Statistical}
    \end{subfigure}
    \begin{subfigure}[b]{0.45\textwidth}
        \centering
        \includegraphics[width=\linewidth]{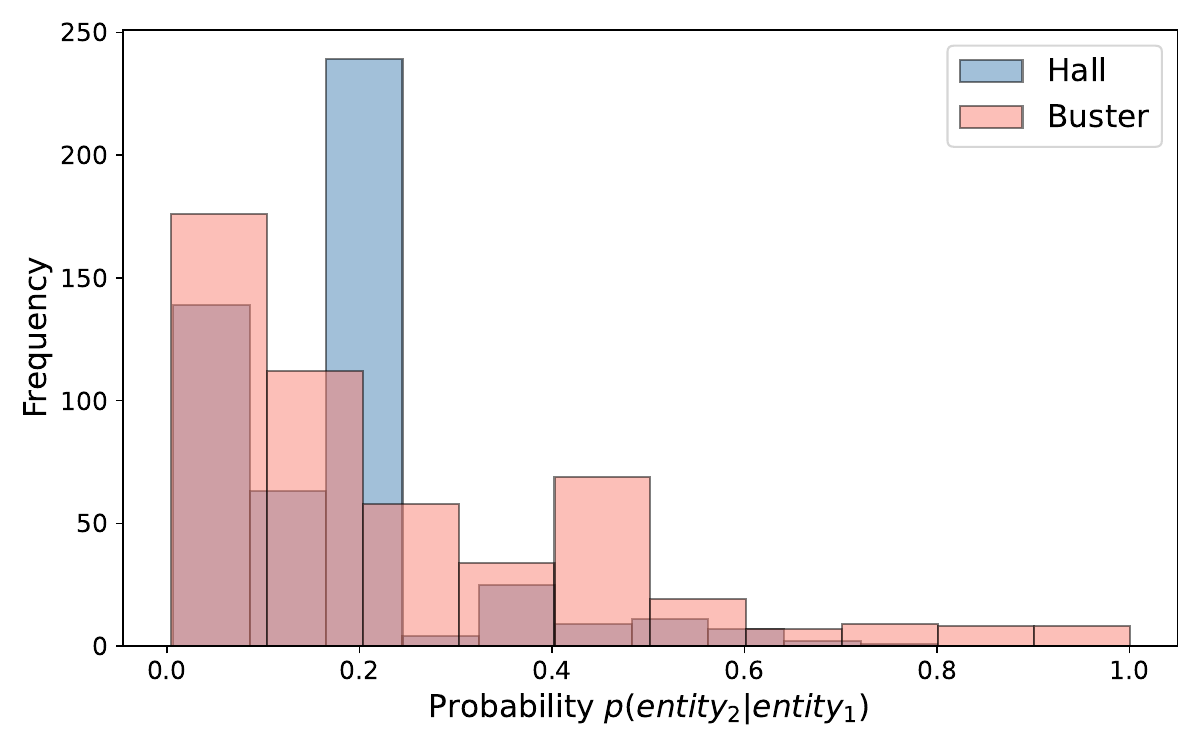}
        \caption{Probabilistic}
    \end{subfigure}   
    \caption{Distribution of the number of objects for statistical insights and probability values $p(\text{entity}_2|\text{entity}_1)$ for probabilistic insights in the created evaluation sets of each dataset.}
    \label{fig:dist}
\end{figure*}

\paragraph{Benchmarking} We present the statistical details of the Hallmarks of Cancer and Buster datasets, along with the extracted triples from each, in Table \ref{tab:stat}. Additionally, Table \ref{tab:stat-eval} provides details of the evaluation sets created for each dataset and insight type. Finally, Figure \ref{fig:dist} illustrates the distribution of the number of objects for statistical insights and the distribution of different probability values $p(\text{entity}_2|\text{entity}_1)$ for probabilistic insights in the evaluation sets for each dataset. To create these evaluation sets, we use the prompt \ref{prompt:extraction} to extract triples of information from documents.

\paragraph{Models} We conduct continual pre-training on LLaMA-3.2 1B, LLaMA-3.2 3B, and LLaMA-3.1 8B models with LoRA and tune hyperparameters on training loss via grid search.
Specifically, tuned hyperparameters include the learning rate $\alpha=[3\times10^{-3}, 10^{-3}, 3\times10^{-4}, 10^{-4}, 3\times10^{-5}, 10^{-5}]$; the LoRA rank $r=[4,8,16]$; the LoRA-alpha $\in \{8,16,32\}$; and the LoRA-dropout $\in \{0.05, 0.1\}$. We trained the LLMs for up to 30 epochs. For probabilistic insights, we set the maximum token length to 200 and sampled the top 10 generated outputs.

\section{Experiments}
\begin{figure*}[th!]
    \centering
    \begin{subfigure}[b]{0.32\textwidth}
        \centering
        \includegraphics[width=\linewidth]{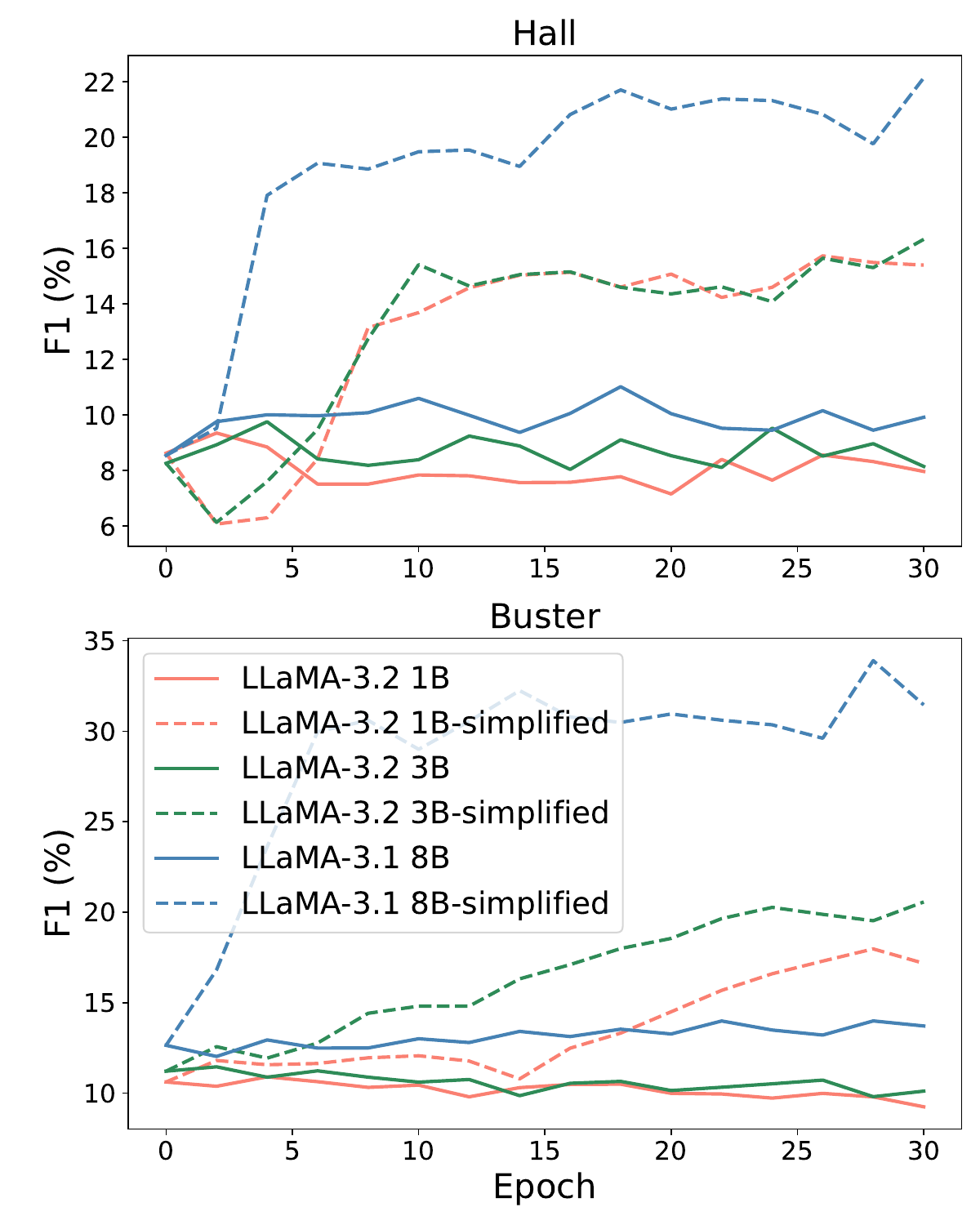}  
        \caption{Declarative}
    \end{subfigure}
    \begin{subfigure}[b]{0.32\textwidth}
        \centering
        \includegraphics[width=\linewidth]{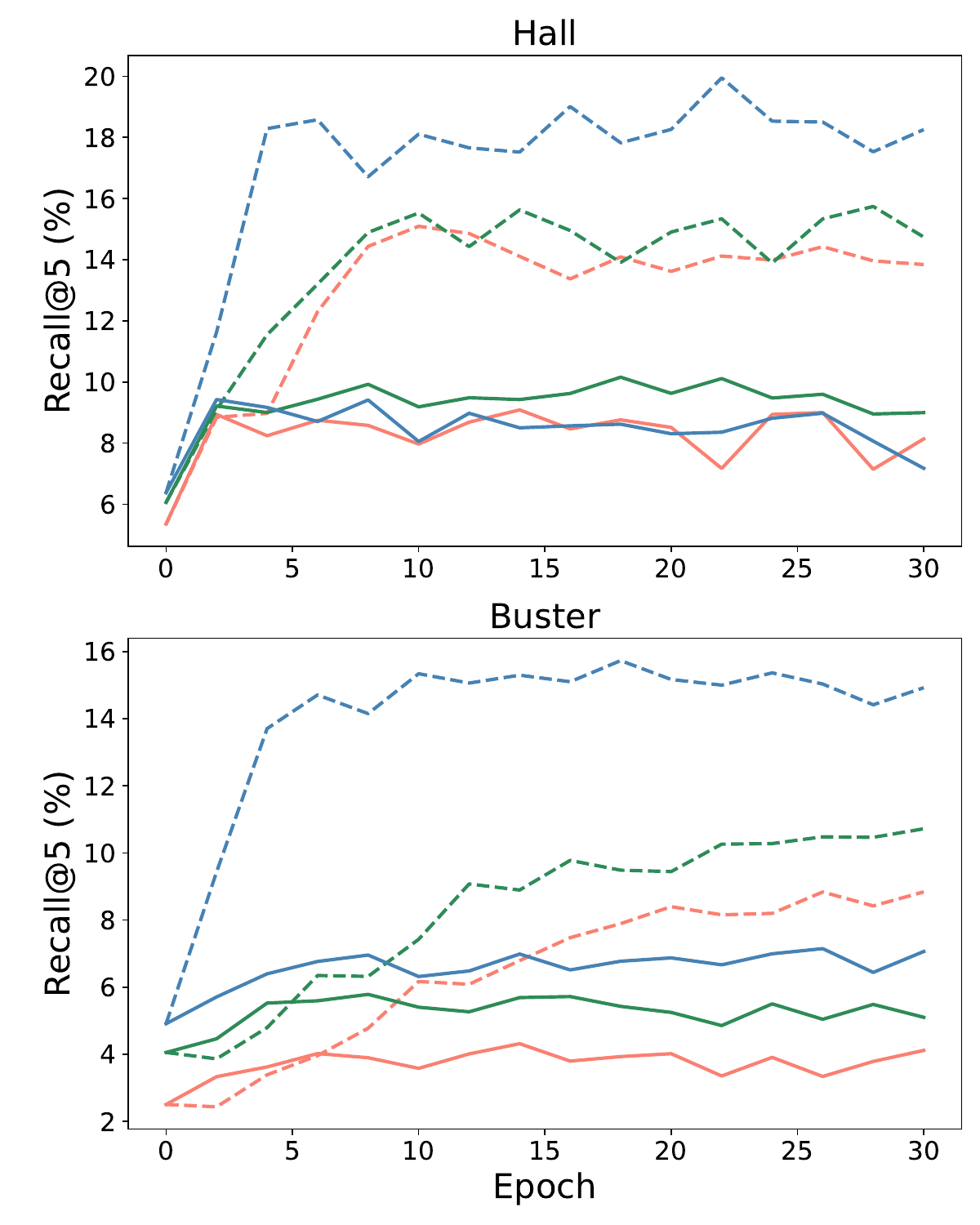}
        \caption{Statistical}
    \end{subfigure}  
    \begin{subfigure}[b]{0.32\textwidth}
        \centering
        \includegraphics[width=\linewidth]{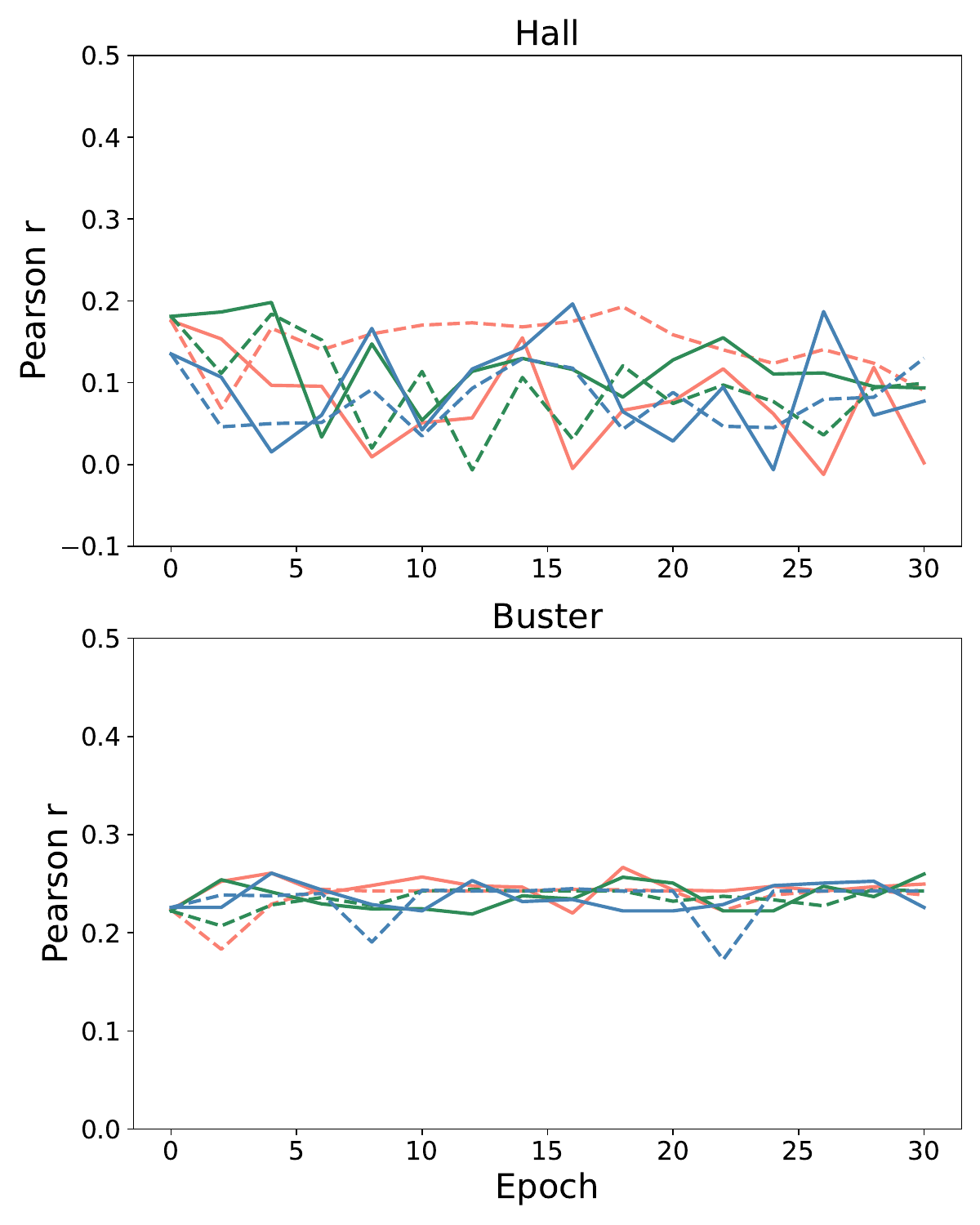}
        \caption{Probabilistic}
    \end{subfigure}  
    \caption{LLM performance on insight extraction during continual pre-training. We report F1 scores for declarative insights, Recall@5 for statistical insights, and Pearson correlation coefficients for probabilistic insights. The results follow similar trends to the metrics in Figure \ref{fig:res}.}
    \label{fig:res2}
\end{figure*}

We provide the F1 score, Recall@5, and Pearson correlation coefficient for declarative, statistical, and probabilistic insights, respectively, in Figure \ref{fig:res2}. The results align with the trends observed in the previously reported metrics. Additionally, Tables \ref{tab:per-rel-hall-dec}, \ref{tab:per-rel-bust-dec}, \ref{tab:per-rel-hall-stat}, and \ref{tab:per-rel-bust-stat} provide a per-relation breakdown of LLM performance for declarative and statistical insights, focusing on the top five most frequent relations. The results show that, despite the top five frequent relations having similar distributions across the data, the impact of continual pre-training and the use of original versus simplified documents on LLM insight-mining performance varies significantly across different relations. This variation may be linked to the differing levels of prior understanding that LLMs have for these relations.

\begin{prompt}[title={\footnotesize\texttt{Triple Extraction Prompt}}, label=prompt:extraction][th!]
Given the following document text, your task is to extract all the triples. The text might have several predicates expressing a relation between a subject and an object.\\
The subject is the entity that takes or undergo the action expressed by the predicate.\\
The object is the entity which is the factual object of the action.\\
The information provided by each predicate can be summarized as a knowledge triplet of the form (subject, relation, object).\\
Extract the information contained in the text in the form of knowledge triplets. \\
Please ignore any non-informative relationships, and focus only on meaningful entities and their relations.\\
Additionally, focus only on information that is self-contained and maintains its meaning independently of surrounding context, ensuring clarity and minimizing ambiguity in relationships. \\
To ensure the triples are self-contained, first extract the information. Then, verify that it does not rely on any preceding or following content. Finally, structure the information into subject, relation, and object format.\\
Each triple should contain a single named entity as the subject and a single named entity as the object. Avoid including multiple entities within either the subject or object.\\
Finally, normalize the relation using your internal knowledge to ensure that relations with the same meaning, but different representations, are mapped to a single, consistent form.\\
Provide the output in JSON format as follows:\\\\

Expected JSON Output:\\
{[\\
  \{\{``subject": ``SUBJECT-1", ``relation": ``RELATION-1", ``object": ``OBJECT-1"\}\},\\
  \{\{``subject": ``SUBJECT-2", ``relation": ``RELATION-2", ``object": ``OBJECT-2"\}\},\\
  ...\\
]}\\\\

Document Text: \{\}
\end{prompt}

\begin{table}[t!]
\small
\centering
\begin{tabular}{llrrr}
\toprule 
&&$\#$ Samples&$\#$ Relations&$\#$ Entities\\
\midrule
\multirow{3}{*}{\rotatebox[origin=c]{90}{\bf Hall}}&Dec&500&103&959\\
&Stat&500&87&2,159\\
&Prob&500&-&548\\
\midrule
\multirow{3}{*}{\rotatebox[origin=c]{90}{\bf Buster}}&Dec&500&162&986\\
&Stat&500&100&3,965\\
&Prob&500&-&586\\
\bottomrule
\end{tabular}
\caption{Data statistics of created evaluation sets.}
\label{tab:stat-eval}
\end{table}

\begin{table*}[t]
\small
\centering
\begin{tabular}{lrrrrrrrrr}
\toprule 
\multirow{2}{*}{\bf Relations} & \multicolumn{3}{c}{\bf LLaMA-3.2 1B}&  \multicolumn{3}{c}{\bf LLaMA-3.2 3B}&  \multicolumn{3}{c}{\bf LLaMA-3.1 8B}\\
\cmidrule(lr){2-4}
\cmidrule(lr){5-7}
\cmidrule(lr){8-10}
&Vanil&CPT-Orig&CPT-Simp&Vanil&CPT-Orig&CPT-Simp&Vanil&CPT-Orig&CPT-Simp\\
\midrule
is a & 1.8&5.4&25.5&1.8&0.0&25.5&1.8&3.6&40.0\\
induced & 3.1&12.5&34.3&3.1&9.4&31.3&3.1&3.1&37.5\\
increases & 5.0&0.0&20.0&5.0&5.0&15.0&5.0&5.0&20.0\\
associated with & 5.9&5.9&29.4&0.0&0.0&35.3&5.9&5.9&35.3\\
causes & 0.0&12.5&18.8&0.0&12.5&31.3&0.0&6.3&50.0\\
inhibits & 0.0&0.0&12.5&0.0&0.0&12.5&0.0&0.0&25.0\\
\bottomrule
\end{tabular}
\caption{Per-relation breakdown of LLMs performance in capturing declarative insights for the top five most frequent relations in the Hallmarks of Cancer dataset. Exact match scores are reported for the vanilla LLMs (Vanil) without training, after 30 epochs of continual pre-training on original documents (CPT-Orig), and after 30 epochs of continual pre-training on simplified documents (CPT-Simp).}
\label{tab:per-rel-hall-dec}
\end{table*}

\begin{table*}[t]
\small
\centering
\begin{tabular}{lrrrrrrrrr}
\toprule 
\multirow{2}{*}{\bf Relations} & \multicolumn{3}{c}{\bf LLaMA-3.2 1B}&  \multicolumn{3}{c}{\bf LLaMA-3.2 3B}&  \multicolumn{3}{c}{\bf LLaMA-3.1 8B}\\
\cmidrule(lr){2-4}
\cmidrule(lr){5-7}
\cmidrule(lr){8-10}
&Vanil&CPT-Orig&CPT-Simp&Vanil&CPT-Orig&CPT-Simp&Vanil&CPT-Orig&CPT-Simp\\
\midrule
provides & 0.0&3.1&0.0&0.0&0.0&3.1&3.1&9.4&50.0\\
is President of & 0.0&0.0&3.2&0.0&0.0&9.7&0.0&0.0&25.8\\
acquired & 0.0&0.0&78.6&0.0&0.0&78.6&3.6&10.7&78.5\\
is approved for & 0.0&0.0&0.0&12.5&0.0&0.0&0.0&12.5&50.0\\
located in & 8.3&8.3&8.3&8.3&8.3&8.3&16.7&16.7&75.0\\
\bottomrule
\end{tabular}
\caption{Per-relation breakdown of LLMs performance in capturing declarative insights for the top five most frequent relations in the Buster dataset. Exact match scores are reported for the vanilla LLMs (Vanil) without training, after 30 epochs of continual pre-training on original documents (CPT-Orig), and after 30 epochs of continual pre-training on simplified documents (CPT-Simp).}
\label{tab:per-rel-bust-dec}
\end{table*}

\begin{table*}[t]
\small
\centering
\begin{tabular}{lrrrrrrrrr}
\toprule 
\multirow{2}{*}{\bf Relations} & \multicolumn{3}{c}{\bf LLaMA-3.2 1B}&  \multicolumn{3}{c}{\bf LLaMA-3.2 3B}&  \multicolumn{3}{c}{\bf LLaMA-3.1 8B}\\
\cmidrule(lr){2-4}
\cmidrule(lr){5-7}
\cmidrule(lr){8-10}
&Vanil&CPT-Orig&CPT-Simp&Vanil&CPT-Orig&CPT-Simp&Vanil&CPT-Orig&CPT-Simp\\
\midrule
is a & 11.6& 16.8&23.5&12.2&18.9&19.5&10.3&15.3&26.2\\
inhibits & 7.3&10.0&15.3&7.2&11.0&10.4&6.6&5.1&19.0\\
induced & 12.3&14.7&23.4&12.3&16.0&24.0&15.5&13.5&30.1\\
increases & 2.0&9.8&7.6&2.8&7.6&7.9&1.9&4.9&16.6\\
decreased & 1.1&3.2&2.6&1.6&4.1&1.1&1.5&2.1&1.0\\
\bottomrule
\end{tabular}
\caption{Per-relation breakdown of LLMs performance in capturing statistical insights for the top five most frequent relations in the Hallmarks of Cancer dataset. Recall@10 are reported for the vanilla LLMs (Vanil) without training, after 30 epochs of continual pre-training on original documents (CPT-Orig), and after 30 epochs of continual pre-training on simplified documents (CPT-Simp)..}
\label{tab:per-rel-hall-stat}
\end{table*}

\begin{table*}[t]
\small
\centering
\begin{tabular}{lrrrrrrrrr}
\toprule 
\multirow{2}{*}{\bf Relations} & \multicolumn{3}{c}{\bf LLaMA-3.2 1B}&  \multicolumn{3}{c}{\bf LLaMA-3.2 3B}&  \multicolumn{3}{c}{\bf LLaMA-3.1 8B}\\
\cmidrule(lr){2-4}
\cmidrule(lr){5-7}
\cmidrule(lr){8-10}
&Vanil&CPT-Orig&CPT-Simp&Vanil&CPT-Orig&CPT-Simp&Vanil&CPT-Orig&CPT-Simp\\
\midrule
acquired & 0.7&1.7&27.1&1.0&2.9&29.4&2.2&9.2&28.1\\
operates & 6.3&12.0&11.0&8.6&11.7&12.1&9.9&12.7&14.9\\
includes & 7.2&6.4&10.6&8.9&12.8&11.9&16.0&16.2&13.8\\
provides & 6.7&6.6&12.0&7.5&10.4&11.0&12.4&14.6&22.3\\
offers & 2.3&7.0&10.5&10.0&10.2&17.3&15.9&18.9&25.7\\
\bottomrule
\end{tabular}
\caption{Per-relation breakdown of LLMs performance in capturing statistical insights for the top five most frequent relations in the Buster dataset. Recall@10 are reported for the vanilla LLMs (Vanil) without training, after 30 epochs of continual pre-training on original documents (CPT-Orig), and after 30 epochs of continual pre-training on simplified documents (CPT-Simp).}
\label{tab:per-rel-bust-stat}
\end{table*}

\end{document}